\documentclass[10pt,twocolumn,letterpaper]{article}

\usepackage{iccv}
\usepackage{times}
\usepackage{epsfig}
\usepackage{graphicx}
\usepackage{amsmath}
\usepackage{amssymb}

\iccvfinalcopy 

\def\iccvPaperID{12} 

\begin{document}

\title{Edge SLAM: Edge Points Based Monocular Visual SLAM}

\author{Soumyadip Maity
\qquad
Arindam Saha
\qquad
Brojeshwar Bhowmick\\
Embedded Systems and Robotics, TCS Research \& Innovation, Kolkata, India\\
{\tt\small \{soumyadip.maity, ari.saha, b.bhowmick\}@tcs.com}
}

\maketitle

\begin{abstract}
   Visual SLAM shows significant progress in recent years due to high attention from vision community but still, challenges remain for low-textured environments. Feature based visual SLAMs do not produce reliable camera and structure estimates due to insufficient features in a low-textured environment. Moreover, existing visual SLAMs produce partial reconstruction when the number of 3D-2D correspondences is insufficient for incremental camera estimation using bundle adjustment. This paper presents Edge SLAM, a feature based monocular visual SLAM which mitigates the above mentioned problems. Our proposed Edge SLAM pipeline detects edge points from images and tracks those using optical flow for point correspondence. We further refine these point correspondences using geometrical relationship among three views. Owing to our edge-point tracking, we use a robust method for two-view initialization for bundle adjustment. Our proposed SLAM also identifies the potential situations where estimating a new camera into the existing reconstruction is becoming unreliable and we adopt a novel method to estimate the new camera reliably using a local optimization technique. We present an extensive evaluation of our proposed SLAM pipeline with most popular open datasets and compare with the state-of-the art. Experimental result indicates that our Edge SLAM is robust and works reliably well for both textured and less-textured environment in comparison to existing state-of-the-art SLAMs.
\end{abstract}

\begin{figure}[t]
\begin{center}
\includegraphics[width=1\linewidth]{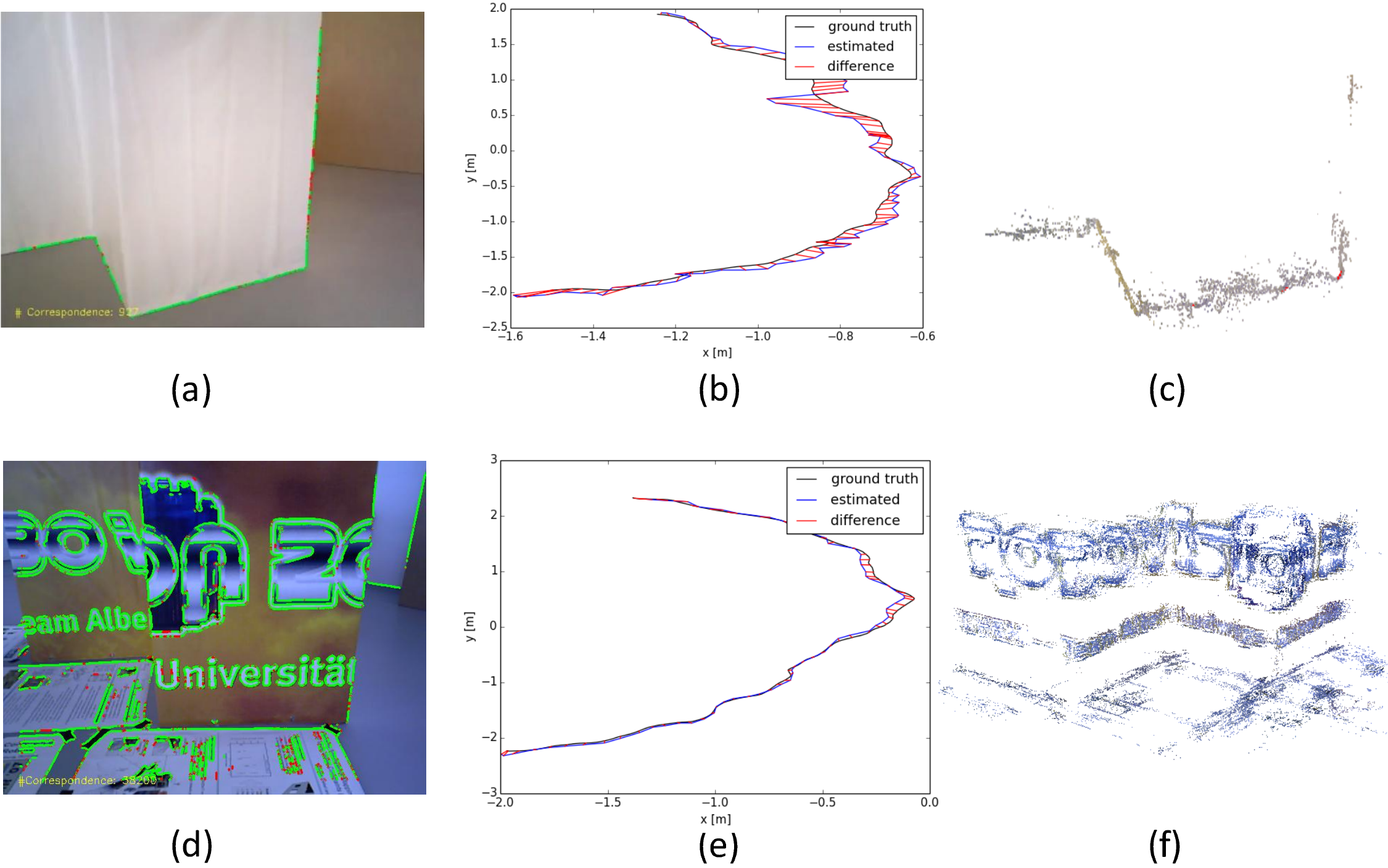}
\end{center}
\caption{(a) A sample image from $fr3\_str\_notex\_far$ \cite{sturm12iros} sequence with edge detection for computing point correspondences. Our Edge SLAM is able to estimate camera poses even on low-textured sequence $fr3\_str\_notex\_far$ \cite{sturm12iros} whereas ORB-SLAM \cite{mur2015orb} is unable to initialize and LSD SLAM \cite{Engel2014} fail to estimate cameras after frame number 357. (b) Comparison of our camera estimates against ground-truth by TUM Benchmarking tool \cite{sturm12iros} for sequence $fr3\_str\_notex\_far$ \cite{sturm12iros}, where the root mean square of absolute trajectory error \cite{sturm12iros} is 6.71 cm. This indicates that our Edge SLAM is reliable for camera track estimation in low-textured environment. (c) Reconstructed sparse 3D structure generated by our Edge SLAM. The structure contain only edge points. (d) A sample image from $fr3\_str\_tex\_far$ \cite{sturm12iros} sequence with extracted edges for point correspondence. (e) Comparison of our camera estimates against ground-truth by TUM Benchmarking tool \cite{sturm12iros} for sequence $fr3\_str\_tex\_far$ \cite{sturm12iros}, where the root mean square of absolute trajectory error \cite{sturm12iros} is 0.65 cm. Our Edge SLAM is able to estimate reliable camera poses in textured sequence also. (f) 3D structure generated by our Edge SLAM on $fr3\_str\_tex\_far$ \cite{sturm12iros} sequence. It shows the accuracy in 3D structure reconstruction in textured environment.}
\label{fig:AdImage}
\end{figure}

\begin{figure}[t]
\begin{center}
\includegraphics[width=1\linewidth]{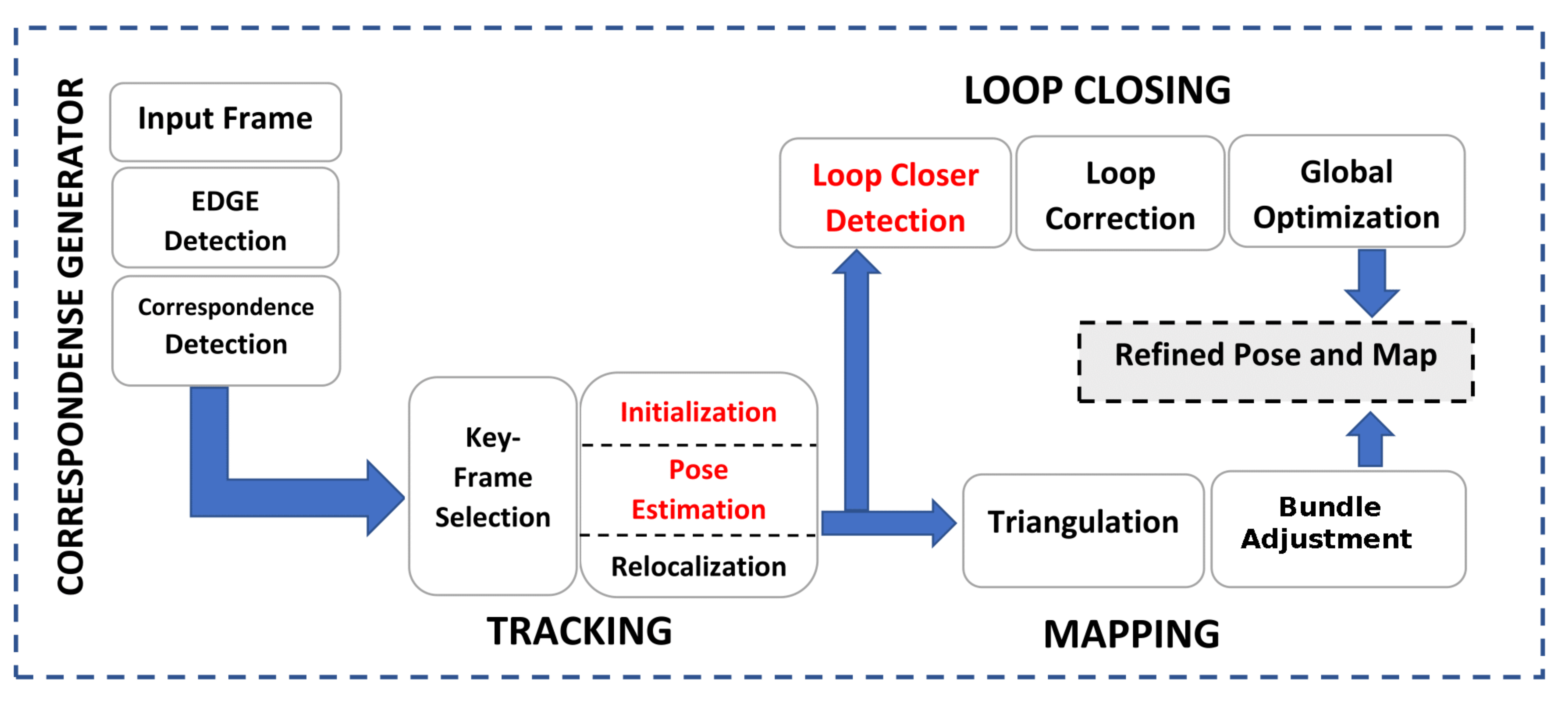}
\end{center}
\caption{Our Edge SLAM system overview, showing major blocks. Contributed blocks are highlighted in red.}
\label{fig:blockDiag}
\end{figure}

\section{Introduction}
\label{sec:intro}
Autonomous navigation of robots requires robust estimation of robot's pose (position, orientation) as well as 3D scene structure. To this end, in recent years, researchers have proposed a variety of algorithms and pipelines for Simultaneous Localization and Mapping (SLAM) \cite{scaramuzza2011visual, DBLP:journals/corr/CadenaCCLSN0L16} using a camera. These visual SLAMs require point correspondences between images for camera (robot) position estimation as well as structure estimation. Feature based methods for visual SLAM \cite{klein07parallel, mur2015orb} try to find the point correspondences between images using SIFT \cite{Lowe:2004:DIF:993451.996342}, SURF\cite{Bay:2008:SRF:1370312.1370556} or ORB features \cite{Rublee:2011:OEA:2355573.2356268}. Using such features, visual SLAMs obtain the camera and structure estimation by minimizing the reprojection error through incremental bundle adjustment \cite{Triggs:1999:BAM:646271.685629}. These SLAMs precisely dependent on the extracted features (SIFT, SURF, ORB) and therefore they miserably fail when the number of points extracted is too less or erroneous especially when the amount of texture present in a scene is very less as shown in Figure~\ref{fig:AdImage}(a) \& (b). Therefore, these SLAMs often produce a partial reconstruction and stop camera tracking when 3D-2D correspondences are less due to insufficient feature correspondences or insufficient 3D points from bundle adjustment. In contrast to feature-based methods, direct methods \cite{Engel2014, endres20143} find such points by minimizing the photometric error \cite{teele1936photometry, walker1974new} and jointly optimizing the poses of the cameras. While these SLAMs are independent of the feature extraction, they are very erroneous in camera pose estimation due to wrong photometric error estimation in lighting change or view change. Moreover, they do not produce good camera estimations in absence of well-textured environment as discussed in Figure~\ref{fig:AdImage}.

In this paper, we use a feature based approach where we detect reliable edges in the image to track the points lying on these edges using bi-directional robust optical flow \cite{Bouguet00pyramidalimplementation}. Such tracking will yield strong point correspondences which are further refined using three-view geometry \cite{Hartley2004}. Using these feature correspondences and epipolar geometry between images we select keyframes for two-view initialization required for structure estimation. Among many such keyframes in an image sequence, we select a particular pair of keyframes using a novel two-view selection method for good initialization. We present a novel two-view selection method for good initialization. We present a comparative result of our novel initialization with existing state-of-the-art methods that clearly exhibit the effectiveness of our method. Then we keep on estimating the new keyframes and the 3D structure using incremental bundle adjustment. Similar to other feature based method, if the 3D-2D point correspondences are ill-conditioned during the addition of a new camera then we apply a novel camera tracking recovery method for continuous tracking of the cameras. If the recovery method fails to produce a reliable camera estimation, the scenario is called track-loss and tries for relocalization. Also, incremental pose estimation accumulates errors introduced at every pose estimation over time resulting in a drift in the camera estimations. Our Edge SLAM uses structural properties of edges in images for computing reliable point correspondences which are used in 3D reconstruction using a local bundle adjustment. We refine the reconstruction globally once a certain number of cameras are estimated. This global bundle adjustment rectifies the drift in the reconstruction. Subsequently, loop closure further refines the camera poses and rectifies such drift. Our Edge SLAM uses structural properties of edges in the images for closing a loop. Our SLAM is robust and reliable in both well-textured and less-textured environment as shown in figure~\ref{fig:AdImage}. The block diagram of our complete SLAM pipeline is shown in figure~\ref{fig:blockDiag} where our contributed blocks are shown in red.

Therefore, the main contributions of this paper are:

\begin{itemize}
\item We use structural properties of edges for correspondence establishment and loop closing.
\item We propose an automatic and robust initialization procedure through validating the reconstructed map quality, which is measured by the geometrical continuation of the map.
\item We propose a novel recovery method of camera tracking in a situation where pose estimation becomes unreliable due to insufficient 3D-2D correspondences. 
\end{itemize}

We organize the remaining part of this paper as follows. In Section~\ref{sec:background}, we review the related work. In Section~\ref{sec:Mathod}, we describe the entire pipeline of our Edge SLAM and evaluate our contribution. Finally, in section~\ref{sec:Result}, we present the experimental results on popular open sequences.
\section{Related Work}
\label{sec:background}
In this section, we describe the related work on SLAM which is broadly divided into feature based methods and direct methods.

{\bf Feature based SLAM:}
Klein and Murray present the first path-breaking visual SLAM, Parallel Tracking And Mapping (PTAM) \cite{klein07parallel} which uses FAST corners points \cite{Rosten:2006:MLH:2094437.2094478} as features and provides simple methods for camera estimation and map generation by decoupling localization and mapping modules. PTAM fails to produce reliable camera estimation in a less-textured environment where availability of point feature is minimal. More recently, Mur-Artal \etal present ORB SLAM \cite{mur2015orb} which uses ORB point feature \cite{Rublee:2011:OEA:2355573.2356268} for point correspondence and yield better accuracy in a well-textured environment. ORB SLAM presents an automatic initialization based on a statistical approach to model selection between planar and non-planar scenes using homography \cite{Hartley2004} or fundamental matrix \cite{Hartley2004} respectively. A better initialization always produces a stable 3D structure, but the reconstructed map has never been used to benchmark initialization because the reconstructed map is sparse. ORB SLAM also fails in challenging low-textured environment. 

{\bf Direct SLAM:}
Direct methods \cite{Newcombe:2011:DDT:2355573.2356447} gains popularity for its semi dense map creation. Recently, Engel \etal present LSD SLAM, \cite{Engel2014} a direct SLAM pipeline that maintains a semi-dense map
and minimizes the photometric error of pixels between images. LSD SLAM \cite{Engel2014} initializes the depth of pixels with a random value of high uncertainty by using inverse depth parametrization \cite{4637878} and optimize the depth based on disparity computed on image pixel. Many times this optimization does not converge with true depth for its noisy initialization and also due to noise present in the computation of photometric errors. Therefore direct methods yield erroneous camera estimation (see table.~\ref{tab:compare}).

{\bf Edge based Visual Odometry:}
Tarrio and Pedre \cite{jose2015realtime} present an edge-based visual odometry pipeline that uses edges as a feature for depth estimation.
But camera estimation is erroneous because odometry works only on pairwise consistency, global consistency checking is very important for accurate camera estimation in a long trajectory. Yang and Scherer \cite{DBLP:journals/corr/YangS17} present a direct odometry based pipeline using points and lines where the estimated camera poses are comparable with ORB SLAM \cite{mur2015orb} for textured environments but the pipeline does not consider the loop-closing which is an integral part of SLAM.

{\bf Point \& Line based SLAM:}
Pumarola \etal \cite{pumarola2017plslam} present PL SLAM, which is built upon ORB SLAM \cite{mur2015orb}. They use line feature along with ORB point feature in tracking and mapping. PL SLAM requires very high processing power for dual feature processing.

Therefore, none of the feature based SLAM or direct SLAM and visual odometry pipelines work reliably well in both well-textured and less-textured environment. In this paper, we try to address this problem by designing an efficient pipeline of feature based SLAM which works in both well-textured and less-textured environments. 

\begin{figure}[t]
\begin{center}
\includegraphics[width=0.8\linewidth]{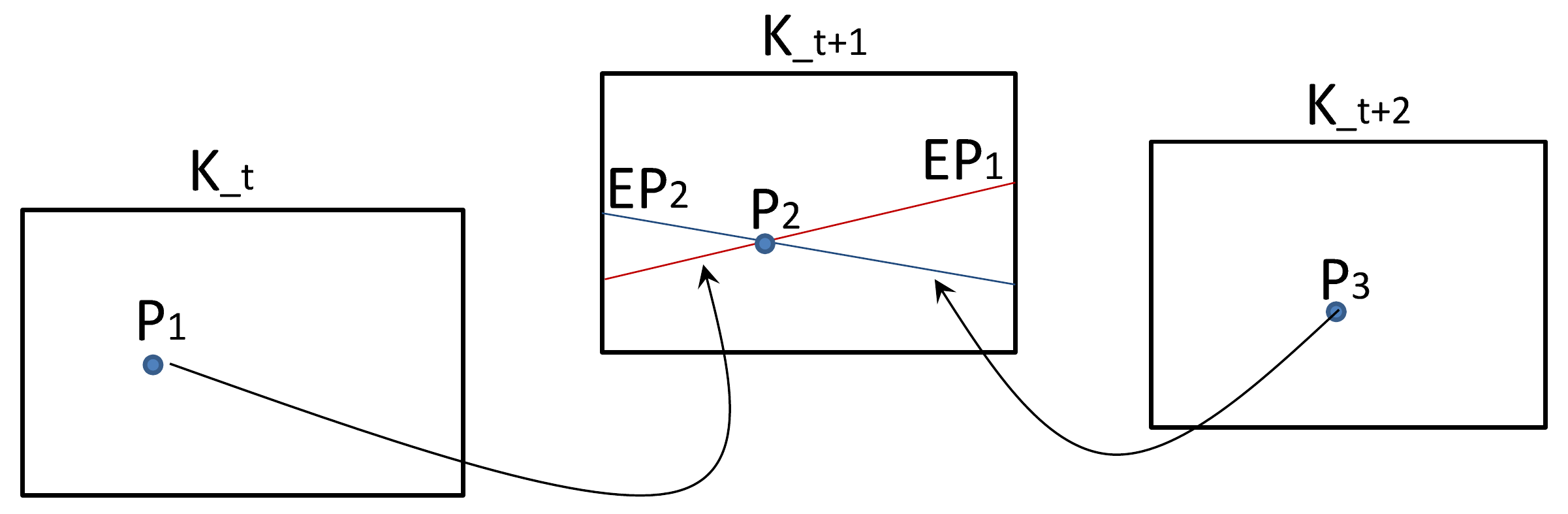}
\end{center}
\caption{Three view validation of correspondence. $EP_1$ and $EP_2$ are the corresponding epilines \cite{Hartley2004} for points $P_1$ \& $P_3$ respectively. If point $P_2$ holds feature correspondence with $P_1$ \& $P_3$, it should be the intersection point of $EP_1$ \& $EP_2$.}
\label{fig:3View}
\end{figure}

\section{Methodology}
\label{sec:Mathod}
\subsection{Correspondence Generation}
\label{subsec:corr}
Feature Correspondence estimation is a major building block which decides the overall performance of any feature correspondence based visual SLAM. Unlike direct methods, we choose a set of effective points to track correspondences. The key idea of feature extraction is to preserve structural information especially edges and therefore we detect reliable edges in an image first. A popular edge detection method is Canny Edge detection \cite{Canny:1986:CAE:11274.11275} which is a gradient based edge detection approach. Edges detected by Canny are not suitable for 2D edge-point tracking as it is sensitive to illumination change and noise. The precise location of an edge and its repetition in consecutive frames is very important for edge-point tracking. We find the DoG based edge detector \cite{Marr187} is reliable due to its robustness in illumination and contrast changes. We thin \cite{conf/dicta/Kovesi10} the DoG edges further to generate edges of a single pixel width. We apply an edge filtering process described by Juan and Sol \cite{jose2015realtime} upon the thinned edges to calculate connectivity of the edge points. This point connectivity information plays an important role in validating edge continuation in different stages of our Edge SLAM pipeline. Edge continuation may not be calculated properly if the input image is blurred or defocused. Those images are high contributing attributes for erroneous feature correspondences as well. We identify and discard those based on an adaptive thresholding method using the variance of gray intensities of edge points of the whole image.
In our Edge SLAM, we estimate feature correspondences of thinned edge points using a bi-directional sparse iterative and pyramidal version of the Lucas-Kanade optical flow \cite{Bouguet00pyramidalimplementation} running on intensity images. We use the window based approach of optical flow to avoid the aperture problem. Point correspondences obtained using only optical flow may contain noise and therefore we remove those noisy correspondences using several filtering methods. We discard the redundant pixels (pixels whose Euclidean distance is very low) first and then remove the points having a higher bi-directional positional error. If a 2D correspondence present in 3 consecutive keyframes (see Section sec.~\ref{subsubsec:key} for keyframe selection), we calculate both the corresponding epilines \cite{Hartley2004} on the middle keyframe as shown in Figure~\ref{fig:3View} and discard 2D correspondences which are not lying at the intersection of the corresponding epilines \cite{Hartley2004}.To reduce the drift, we remove the forward correspondences, which are non edge points.

\begin{figure}[t]
\begin{center}
\includegraphics[width=0.7\linewidth]{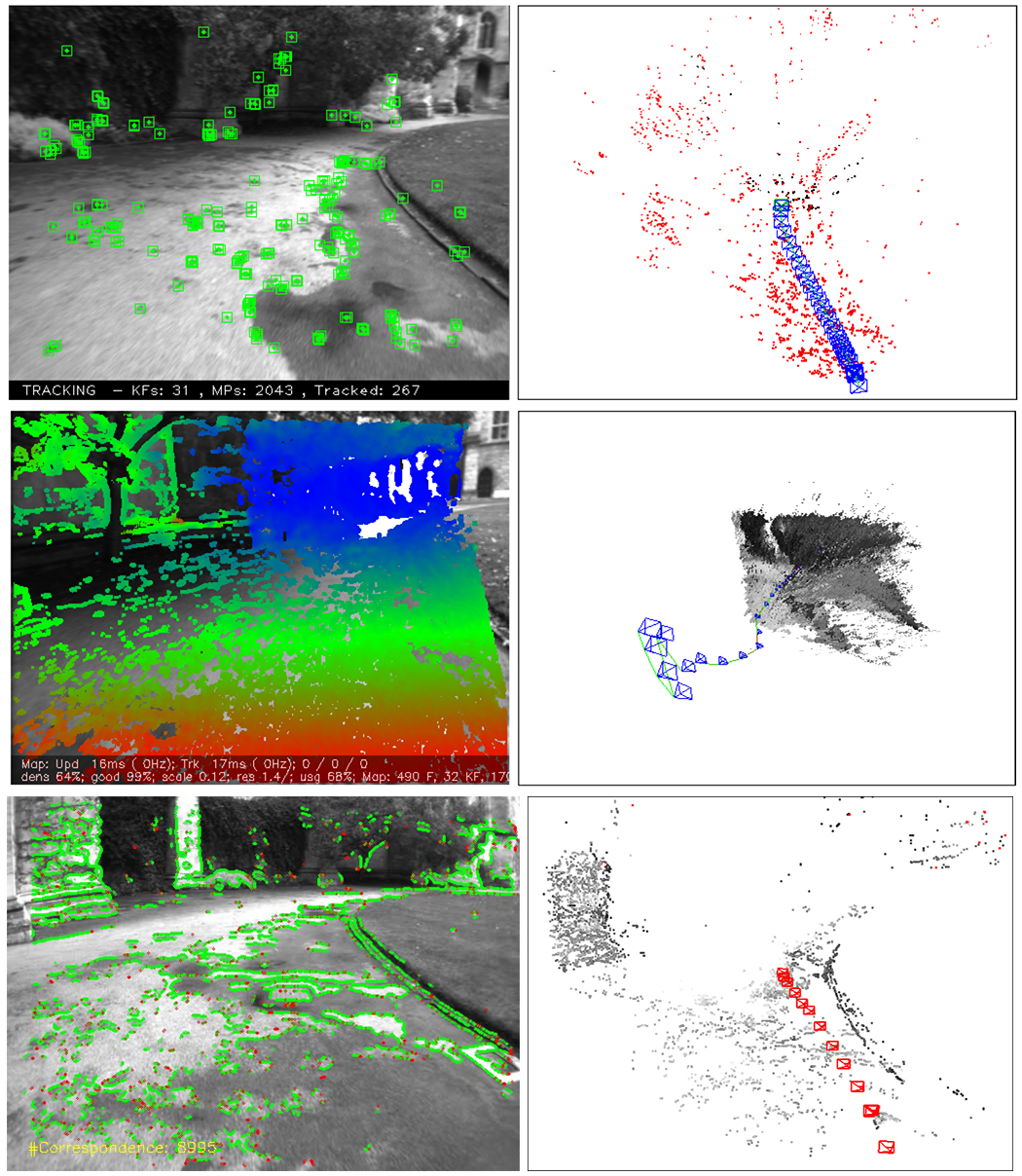}
\end{center}
\caption{Few frames after initialization in the NewCollege sequence \cite{smith2009new}. Top: ORB-SLAM, initialize camera poses calculating fundamental matrix but the reconstructed 3D structure (red points) does not convey any meaning full understanding of the scene. Middle LSD-SLAM, initialize the map with erroneous planar depth. Bottom: Edge-SLAM, Initialize camera poses with a continuous 3D structure. Continuous edges (\eg wall on left side in the point cloud, separation line on the right side in the point cloud) are visible on the picture.}
\label{fig:initCompare}
\end{figure}

\begin{figure}
\begin{center}
\includegraphics[width=1\linewidth]{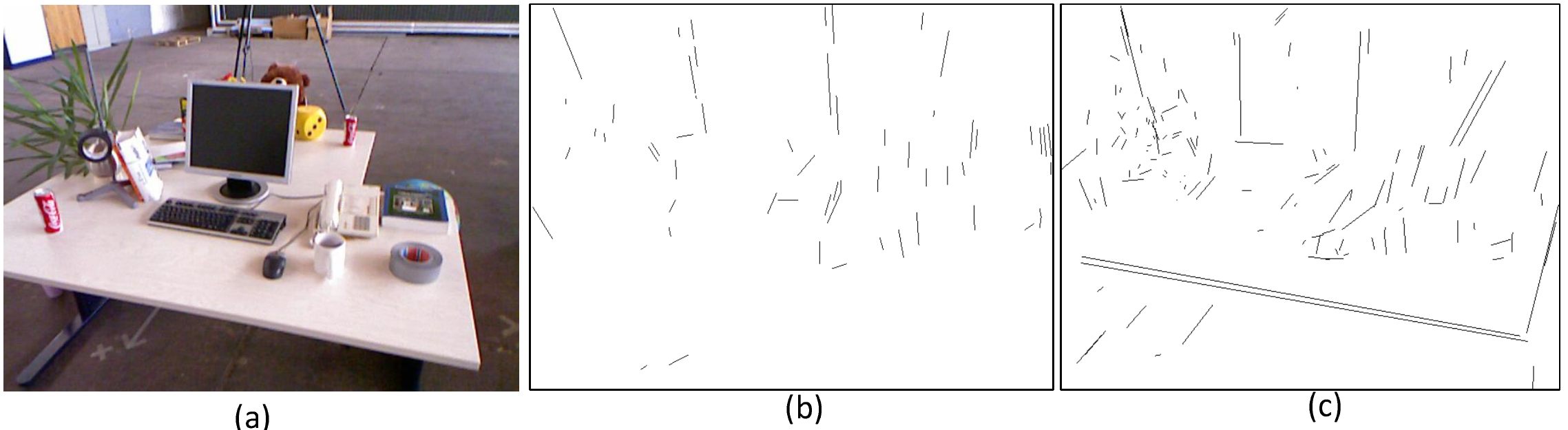}
\end{center}
\caption{(a) A sample image from $fr2\_desk$ \cite{sturm12iros} sequence. (b) Initialization at frame number 11 reconstructs discontinuous 3D structure. Due to low quality-factor, our pipeline discard this initialization. If the pipeline accepts the initialization and continues, at frame number 117 it produces 1.12 cm root mean square of absolute trajectory error \cite{sturm12iros} against ground-truth. (c) Initialization at frame number 34 generates continuous 3D structure. Due to higher quality-factor our pipeline select it as valid initialization and at frame number 117 it produces 0.48 cm root mean square of absolute trajectory error \cite{sturm12iros} against ground-truth.}
\label{fig:init}
\end{figure}

\subsection{Keyframe Selection}
\label{subsubsec:key}
Keyframes are a subset of frames, which we choose to avoid low parallax error and redundancy for robotic movement. The overall accuracy of the system varies on keyframe selection and number of keyframes. ORB SLAM [17] presents usefulness of keyframe selection and we adopt a similar technique which is suitable for our pipeline. Let $I_{t-1}$ and ($I_t$) are two consecutive frames. We process current frame ($I_t$) and last selected keyframe  ($K_m$), where $0 \leq m < t-1$, for next keyframe selection. We select next keyframe $K_{m+1}$ using following criteria if any one of the following conditions holds:
\begin{itemize}
\item We calculate pairwise rotation between $I_t$ and $K_m$ using epipolar geometry \cite{Nister:2004:ESF:987526.987623}. If the rotation is more than $15^{\circ}$ then $I_t$ is not reliable as optical-flow may produce noisy correspondences. We consider ($I_{t-1}$) as the new keyframe ($K_{m+1}$).
\item We compute the average number of points tracked as correspondences for every image. If the number of 2D feature correspondences between $I_t$ and $K_m$ reduced below thirty percent of the average feature correspondences then $I_t$ is not reliable as there may be a sudden scene or illumination change and we consider ($I_{t-1}$) as the new keyframe ($K_{m+1}$).
\item If number of 3D-2D correspondences reduces below 250, we consider ($I_{t-1}$) as a new keyframe ($K_{m+1}$).
\item We compute the average positional change of feature correspondences between $I_t$ and $K_m$ by averaging  Euclidean distance between previous and current pixel positions of all correspondences. If average positional change is more than twenty percent of the image width, we consider current frame ($I_t$) as a new keyframe ($K_{m+1}$).
\item If none of the previous conditions occur, we consider new keyframe ($K_{m+1}$) in a fixed interval of 1 second.
\end{itemize}

\subsection{Two-view Initialization}
\label{subsubsec:twoviewresec}
SLAM, being an incremental camera estimation pipeline, uses incremental bundle adjustment for estimating cameras. To initialize the bundle adjustment many of the SLAM techniques select two-view as seed pair for the initialization of the cameras after computing epipolar geometry \cite{Nister:2004:ESF:987526.987623} between the image pair and triangulate \cite{Hartley1995} the point correspondences for initial 3D structure followed by refinement using bundle adjustment. Then new cameras are added into the existing reconstruction through re-sectioning utilizing 3D-2D correspondences. Therefore, the seed-pair used for initialization of bundle adjustment plays an important role to determine the quality of structure estimation which in turn produce correct camera trajectory through re-sectioning. However, such camera and structure estimation may not be well constrained under low-parallax and the output may suffer from ambiguity and drift \cite{longuet1986reconstruction}. Therefore, many of the existing SLAMs use geometrical validation of camera poses for reliable initialization of cameras and structure \cite{klein07parallel, mur2015orb}.

In our Edge SLAM, we also use incremental bundle adjustment. We choose two keyframes based on any one of the following conditions hold:
\begin{itemize}
\item Pairwise rotation between the keyframes is more than $15^{\circ}$.
\item Averaging Euclidean distance between previous and current pixel exceed 20 percent of input image width.
\item Time difference of 1 second time between the frames.
\end{itemize}

We generate the initial 3D structure based on initial pairwise pose estimation by Five-point algorithm from Nister \cite{Nister:2004:ESF:987526.987623} followed by triangulation \cite{Hartley1995}. We further optimize the initial structure using Bundle Adjustment. For correctness of this initialization we introduce a novel map validation method. We find the spatial smoothness among 2D points which should also conform to the reconstructed 3D points. An edge can be considered as connected straight lines of small length and a straight line in 3D should have same continuation. The ordering of points in straight lines should remain same in both 2D and 3D. In particular, any two neighbouring edge points in 2D should also be neighbour in 3D. We identify those straight lines based on local neighbourhood gradient. The number of such corresponding small straight lines in both 2D images and in reconstructed point cloud using two-views signify the quality of the map. We assign a quality-factor to the map based on that number and keep on finding the two views until the quality-factor is above a threshold. If the quality-factor is greater than a threshold, we fix those seed pair as a valid initialization of bundle adjustment and keep tracking of cameras using new views.

Figure~\ref{fig:init} shows an example of our initialization on dataset $fr2\_desk$ which is a TUM RGB-D benchmark \cite{sturm12iros} dataset. Initialization until frame number 11 produces discontinuous structure see figure~\ref{fig:init}(b) for a sample two-view initialization using keyframes within first 11 frames. The root mean square (RMS) of Absolute Trajectory Error (ATE) ~\cite{sturm12iros} against ground-truth till frame number 117 with such initialization is 1.12 cm. Our pipeline rejects this initialization for discontinuity in reconstructed structure. Instead, we initialize from frame number 34 (suggested by our quality metric and the corresponding initial structure is shown in figure~\ref{fig:init}(c)) using which the RMS of ATE \cite{sturm12iros} till frame number 117 is 0.48 cm. This clearly shows that our two-view selection strategy using the quality of the initial structure is significantly better than initialization using any two views which produce very sparse or noisy 3D structure. Apart from the improvement in camera pose, our initialization also produces a better 3D structure for SLAM. Figure 4 shows an example where due to our good two-view initialization the 3D structure obtained after a certain period of time by our pipeline is significantly better than LSD-SLAM \cite{Engel2014} or ORB SLAM \cite{mur2015orb}. We attribute the failures of LSD-SLAM and ORB-SLAM for producing good structure is due to the initialization using a planner depth and initialization without structure validation respectively.

\subsection{Tracking and Mapping}
\subsubsection{Incremental Pose Estimation \& Mapping}
\label{subsubsec:IncrePose}

Using the initial 3D structure from two-view bundle adjustment, we keep on adding new cameras through re-sectioning \cite{Epnp09}. We only add new keyframe into the existing reconstruction instead of all frames.
Our re-section based pose estimation method for keyframe only is more general than constant velocity motion model used by ORB SLAM  \cite{mur2015orb} using every frame. The accuracy of initial estimation by constant velocity motion model may drastically fall for any non-linear motion.

\begin{figure}[t]
\begin{center}
\includegraphics[width=1\linewidth]{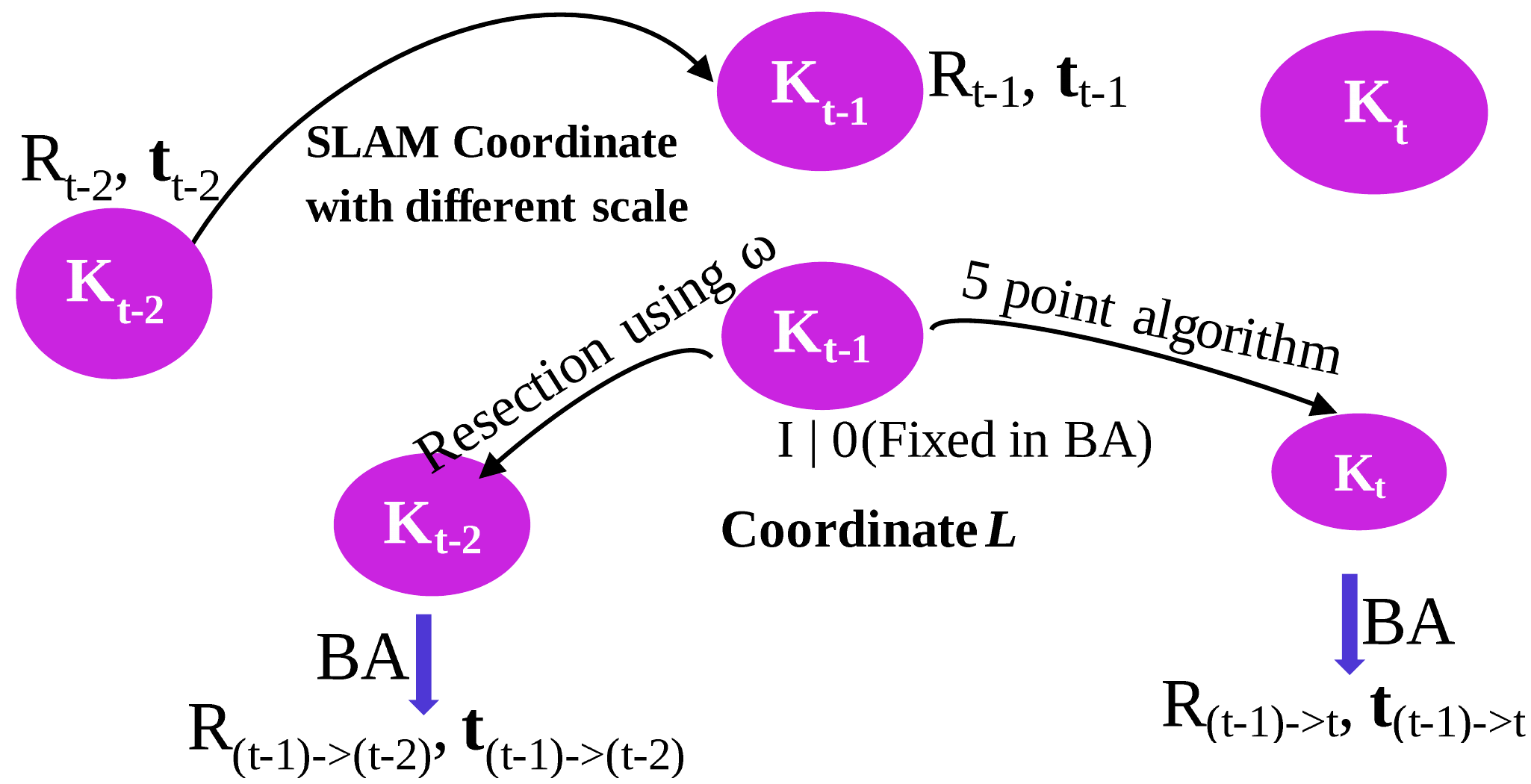}
\end{center}
\caption{Track-loss recovery: Images $K_{t-2}$, $K_{t-1}$ and $K_t$ are estimated separately in $L$ coordinate and these estimations are transformed to SLAM coordinate by coordinate merging after scale correction.}
\label{fig:track-loss}
\end{figure}

After two-view initialization we find the next keyframes using keyframe selection method described in Section~\ref{subsubsec:key}. We add these keyframes into the existing reconstruction by re-sectioning using 3D-2D correspondences followed by addition of new structure points through triangulation \cite{Hartley1995}. We define two keyframes as co-visible keyframes, only if they have more than 100 common visible 3D points. We choose $C\subseteq$ $K$, where $K$ is the set of all keyframes, $C$ contains the new keyframe $K_m$ and its co-visible keyframes. 3D point set $B$ contains all 3D points visible by all the keyframes of $C$. Now we proceed local bundle adjustment on $C$ and $B$.

\begin{equation}
\min_{ C_j,B_i}\sum_{i=1}^{n}\sum_{j=1}^{m} V_{ij}D(P( C_j, B_i),\textbf{x}_{ij}\Psi(\textbf{x}_{ij}))\label{bundle1}
\end{equation}

where, $V_{ij} \in \{0,1\} $ is the visibility of the $i^{th}$ 3D point in the $j^{th}$ camera, $P$ is the function which projects a 3D point $ B_{i}$ onto camera $ C_{j}$ which is modelled using $7$ parameters ($1$ for focal length, $3$ for rotation, $3$ for position) , $\textbf{{x}}_{ij}$ is the actual projection of the $i^{th}$ point onto the $j^{th}$ camera, $\Psi(\textbf{{x}}_{ij}) = 1 + r\|\textbf{{x}}_{ij}\|^2$ is the radial distortion function with a single parameter ($r$) and $D$ is the Euclidean distance.  We minimize  (equation~\ref{bundle1}) using \cite{Triggs0}. We fix the focal length of the camera using the known internal calibration and pose of the cameras as well as map points are optimized.

We refine the pose and reconstruction globally using global bundle adjustment in every 25 second or 25 locally optimized keyframe accumulated. This refinement using global bundle adjustment correct the drift, if any, in the estimated camera trajectory.

\subsubsection{Track-loss Handling}
\label{subsec:trackloss}
Track-loss is one of the major issues present in the literature of visual SLAM where estimated camera track and reconstructed map breaks due to tracking failure during re-section. Some of the dominant reasons for such failure are occlusion, abrupt motion, corresponding structure point reduction in bundle adjustment, reduction in the number of 2D point correspondences due to low-textured environment etc. Every existing SLAM tries to relocalize after track-loss but relocalization is not guaranteed in robotic movements unless the camera returns back very near to its previously visited location. To mitigate the track-loss problem without relocalization, we take a sequence of three consecutive keyframes $K_{t-2}$, $K_{t-1}$ and $K_t$ where first two keyframes are already estimated in SLAM coordinate and we are unable to track current keyframe $K_t$ using re-section. There are two possible reasons for such failures in re-section.

{\bf Case 1:} Here 2D correspondences exist in all three keyframes $K_{t-2}$, $K_{t-1}$ and $K_t$ but our pipeline rejects the corresponding 3D-2D correspondences due to high reprojection error on keyframes $K_{t-2}$ and $K_{t-1}$. So here we try to use these 2D correspondences for estimating $K_t$.

{\bf Case 2:} Here 2D correspondence exists till keyframe $K_{t-1}$ and absent in $K_t$ due to unstable optical flow. Therefore 3D-2D correspondences are in ill-conditioned for keyframe $K_t$. We keep this case out of the scope as optical flow fails.

We try to estimate the keyframe $K_t$ separately from SLAM coordinate using 2D-2D point correspondences. Epipolar geometry \cite{Nister:2004:ESF:987526.987623} provides a pairwise rotation and an unit direction from frame $K_{t-1}$ to frame $K_t$ in a coordinate system where frame $K_{t-1}$ is at origin. We name this coordinate system as $L \in {\rm I\!R}^3$ in rest of the paper, which has an unknown scale difference with SLAM coordinate system. We create new 3D points with 2D correspondences and estimated poses of frames $K_{t-1}$ and $K_t$ in $L$ through triangulation \cite{Hartley1995} and re-estimate the frame $K_{t-2}$ in $L$ by re-sectioning utilizing 3D-2D correspondences. Bundle adjustment further refines the poses for frames $K_{t-2}$ and $K_t$ without changing the coordinate system \ie pose for frame $K_{t-1}$ remain unchanged after bundle adjustment. Bundle adjustment produces stable pairwise rotation and translation pair ($R_{(t-1)\rightarrow t}$, ${\bf t}_{(t-1)\rightarrow t}$), ($R_{(t-1)\rightarrow (t-2)}$, ${\bf t}_{(t-1)\rightarrow (t-2)}$) for frames $K_t$ and $K_{t-2}$ respectively. The pipeline continues only when bundle adjustment produces enough inliers 3D points (more than 100) otherwise try for relocalization. The main idea is to merge the coordinate $L$ with the SLAM coordinate after correcting the scale difference. Figure~\ref{fig:track-loss} represents the method graphically where three connected frames $K_{t-2}$, $K_{t-1}$, $K_t$ are estimated in $L$ and then merge with SLAM coordinate.

We calculate the scale difference between two coordinate systems using the estimations of frames $K_{t-2}$ and $K_{t-1}$ in both the SLAM and $L$ coordinate systems using the equation~\ref{equ:scaleCalc} where $C_{Li}$  and $C_{Si}$ denote the camera center of $i^{th}$ frame in $L$ and SLAM coordinate system respectively. Scale corrected camera center $C_{scaled\_t}$ for frame $K_t$ in $L$ follow the relation as given in equation~\ref{equ:scaleRec}.

\begin{table}[t]
\begin{center}
\resizebox{\columnwidth}{!}{%
\begin{tabular}{|l c c| c c|}
\hline
 Sequences & ORB-SLAM & LSD-SLAM & No. of times & RMS of\\
 & Track lost & Track lost & track-loss & ATE (cm)\\
 & & & recovery is used & \\
 \hline\hline
 $fr3\_str\_notex\_near$ & Unable to & frame 533 & 12 & 8.29\\
 & initialize & & & \\
 \hline
 $fr3\_str\_notex\_far$ & Unable to & frame 357 & 7 & 6.71\\
 & initialize & & & \\
 \hline
 $ICL/office1$ & frame 354 & frame 193 & 5 & 19.5 \\
 \hline
\end{tabular}}
\end{center}
\caption{Comparison among ORB SLAM, LSD SLAM and our Edge SLAM in some low-textured sequences along with the details of our track-loss recovery method.}
\label{tab:M2compare}
\end{table}

\begin{figure}
\begin{center}
\includegraphics[width=1\linewidth]{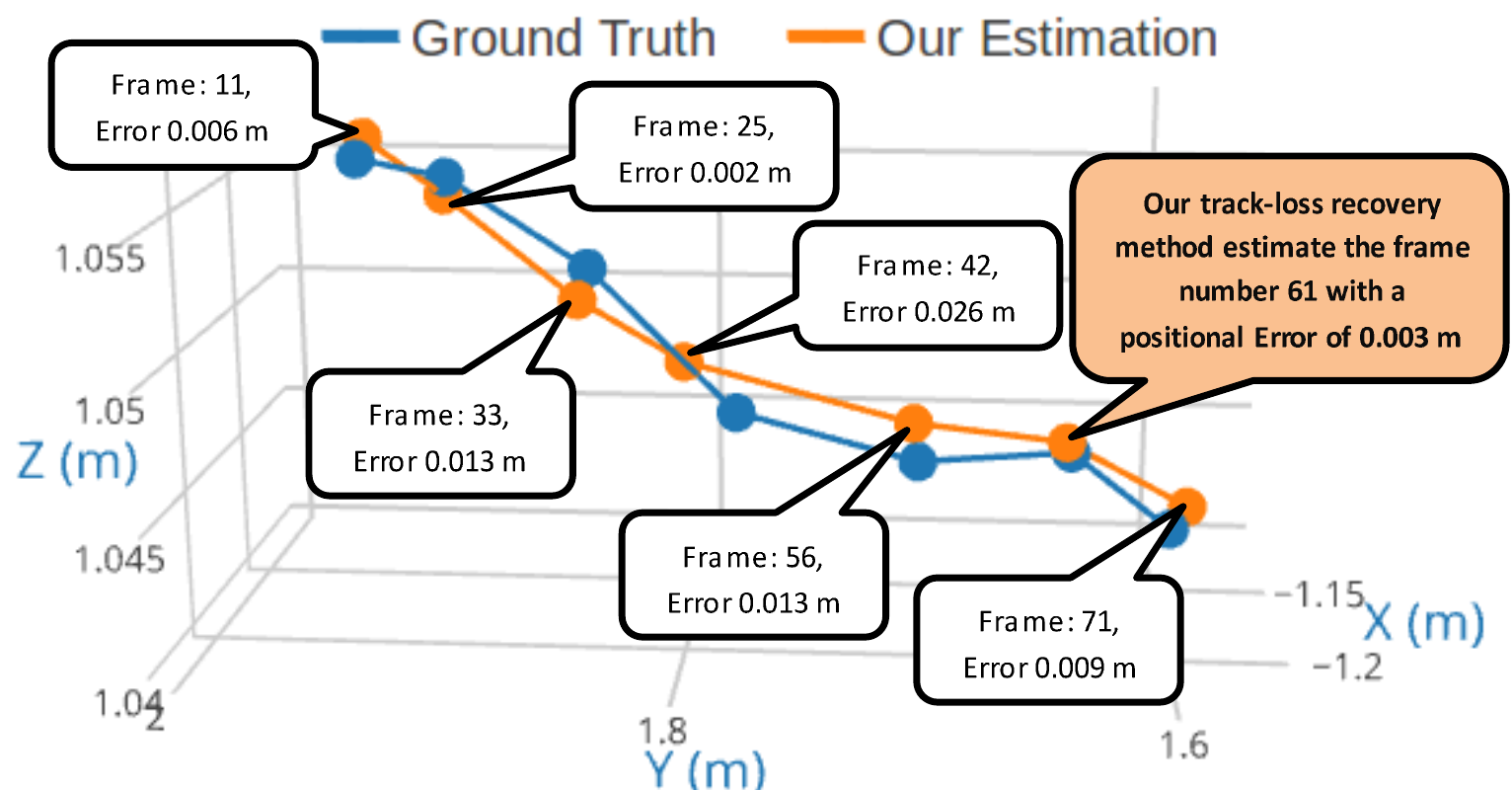}
\end{center}
\caption{Validation of track-loss avoidance in $fr3\_str\_notex\_far$ \cite{sturm12iros} sequence. Camera estimation become unreliable as 3D-2D correspondences become very low at frame number 61. Our track-loss recovery method estimates the frame with 0.3 cm absolute positional error against ground-truth.}
\label{fig:M2_Survival}
\end{figure}

\begin{equation}
{\lambda}_{init} = \frac{C_{S(t-1)} - C_{S(t-2)}}{C_{L(t-1)} - C_{L(t-2)}}
\label{equ:scaleCalc}
\end{equation}

\begin{equation}
C_{scaled\_t} = {\lambda}_{init} (-R_{(t-1)\rightarrow t}^T {\bf t}_{(t-1)\rightarrow t})
\label{equ:scaleRec}
\end{equation}

We require to align the axes of both the coordinate system in order to merge them. Therefore, we rotate the SLAM coordinate axes to aligned with frame $K_{t-1}$ and calculate the camera center for frame $K_t$ in the rotated SLAM coordinate through equation~\ref{equ:camCenterAddition} and calculate the camera center for frame $K_t$ in the SLAM coordinate through a reverse rotation as given in equation ~\ref{equ:camCenterSLAM}.

\begin{equation}
C_{S\_t\_rot}^T = C_{S(t-1)}^T R_{t-1}^T + C_{scaled\_t}^T
\label{equ:camCenterAddition}
\end{equation}

\begin{equation}
C_{S\_t}^T = C_{S\_t\_rot}^T * R_{t-1}
\label{equ:camCenterSLAM}
\end{equation}

Pairwise rotation is always independent of any coordinate system and thus we use the pairwise rotation of frame $K_t$ to get absolute rotation $R_t$ in SLAM coordinate using equation~\ref{equ:pairwiseRotation} \cite{Hartley2004} where $R_{t-1}$ is the absolute rotation of the frame $K_{t-1}$ in SLAM coordinate. 

\begin{equation}
R_{(t-1)\rightarrow t} = R_t * R_{t-1}^T
\label{equ:pairwiseRotation}
\end{equation}

Finally we calculate the translation vector (${\bf t}_{S\_t}$) of frame $K_t$ in SLAM coordinate using equation~\ref{equ:trans} and include the 2D correspondences present in $\omega$ set to SLAM coordinate by triangulation \cite{Hartley1995} for better connectivity between current frame with previous frames. Estimated pose for the frame $K_t$ in SLAM coordinate is little erroneous and local bundle adjustment further refine the poses. If the bundle adjustment produces enough inliers, incremental tracking procedure continues from next frame onwards otherwise the pipeline initiate entire track-loss avoidance procedure from beginning for next frame.

\begin{equation}
{\bf t}_{S\_t} = -R_t * C_{S\_t}
\label{equ:trans}
\end{equation}

We evaluate our track-loss recovery method with very less-textured data, for \eg sequence $fr3\_str\_notex\_far$~\cite{sturm12iros} where camera estimation become unreliable at frame number 61 due to insufficient 3D-2D correspondences. Our track-loss recovery method estimates the keyframe with a positional error of 0.3 cm against ground-truth. Figure~\ref{fig:M2_Survival} shows the result of the given instance.

Table~\ref{tab:M2compare} presents the detail result of our track-loss recovery method on some standard sequences of TUM RGB-D benchmark \cite{sturm12iros} and ICL-NUIM \cite{handa:etal:ICRA2014}. The result (in Table~\ref{tab:M2compare}) shows ORB SLAM is unable to initialize in first 2 sequences and camera tracking is failed in the ICL/office1. LSD SLAM failed in camera tracking in all the sequences. Our track-loss recovery method produce correct estimation of camera poses in all such situations (details are given in Table~\ref{tab:M2compare}).

\begin{figure}
\begin{center}
\includegraphics[width=1\linewidth]{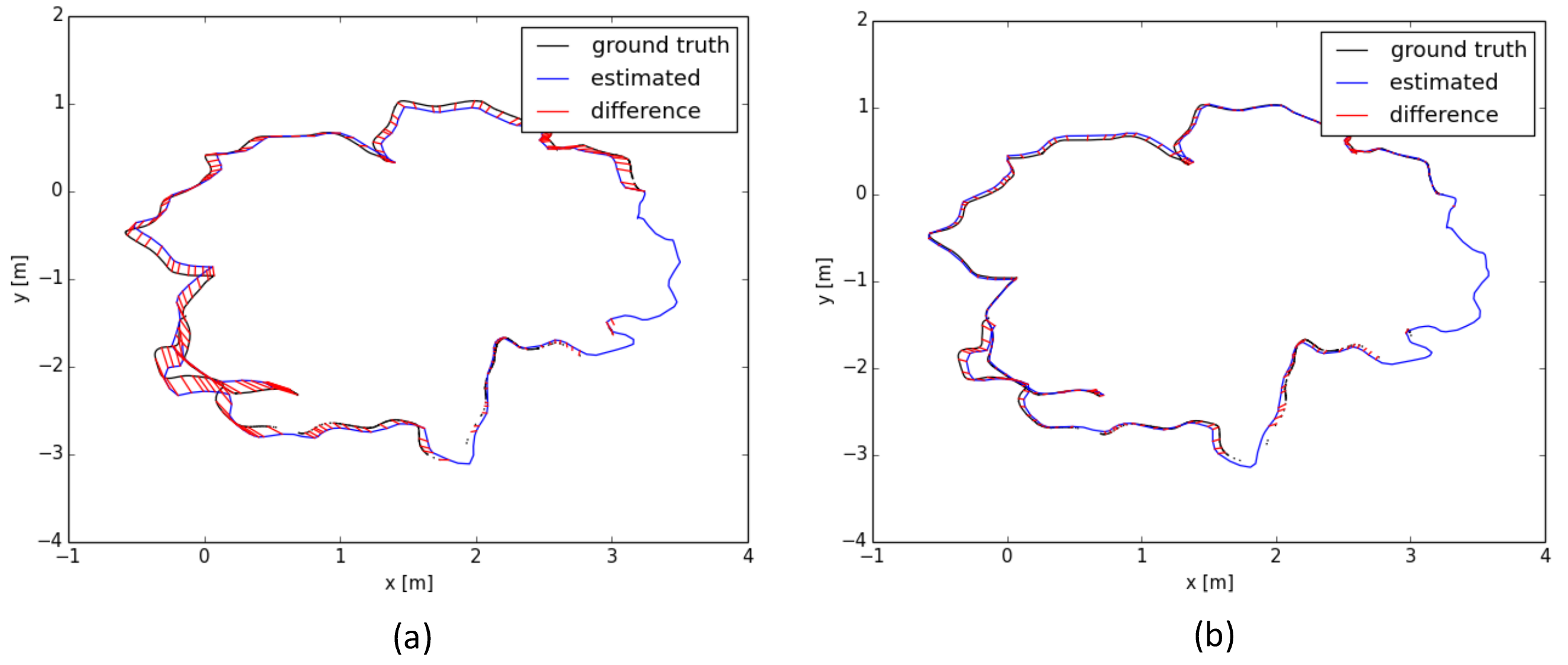}
\end{center}
\caption{Estimated camera trajectory comparison on $fr2\_xyz$ \cite{sturm12iros} sequence. (a) Before loop closing. The RMS of ATE is 9.5 cm. (b) After loop closing. The RMS of ATE is 1.75 cm}
\label{fig:loop}
\end{figure}

\subsection{Loop Closing}
\label{sec:loop}
Incremental bundle adjustment estimates the cameras incrementally after two-view initialization. Such incremental pose estimations accumulate errors and create a drift in estimated trajectory. Loop closure tries to rectify such drift by matching structural properties of images between non-neighbouring keyframes (does not share any point correspondence). 

\subsubsection{Loop Detection}
\label{sec:loopdetect}
We consider loop if two non-neighbouring estimated keyframes share a similar scene from a different view point. Our loop detection method tries to find loop only on keyframes. A reliable loop detection method should be invariant to scale changes, translation, rotation of scene structure. Image moment invariants \cite{Hu1962} are invariant of those specific transformations and are very good features to use when dealing with a large set of images and a large displacement. This is a well-accepted technique to classify objects \cite{keyes2001using} as well as for pattern recognition \cite{Flusser:2009:MMI:1708062} where it identifies objects through edge matching by the third order moments invariants of polygon \cite{Hu1962}. So we exploit image moment invariants and match edges of keyframes based on third order moments \cite{Hu1962}. Subsequently, we adopt a multilevel matching mechanism where every keyframe is divided into 16 quadrants and matching score is calculated based on the matched quadrants between a pair of keyframes. An adaptive weighted average of the number of edges, average edge pixel density and average intensity in each quadrant are used to find the matched quadrant. We derive a final matching score between two keyframes, averaging the matching score of moments invariants and voting. Initially, our method calculates matching score of last keyframe $K_t$ with maximum 5 previous keyframes, which have less than $30^{\circ}$ degree viewing direction change (immediate neighbour with high cohesion) and retain the lowest matching score $M_{min}$ as the threshold to choose matching with non-neighbouring keyframes. Subsequently, it calculates the matching score of all non-neighbouring keyframes with $K_t$ and retains only the keyframes that are having a matching score more than $M_{min}$. We consider a keyframe as a matched keyframe with $K_t$ only if three consecutive keyframes are having matching score more than $M_{min}$ to avoid wrong matches. Our method select the keyframe as loop candidate ($K_{loop}$) having maximum matching score among the three consecutive keyframes. Subsequently, we calculate point correspondences between keyframe $K_t$ and $K_{loop}$ as described in sec.~\ref{subsec:corr}. These point correspondences create a set of 3D-3D correspondences between $K_t$ and $K_{loop}$. We calculate a similarity transformation $T_{sim}$ between $K_t$ and $K_{loop}$ using these 3D-3D correspondences as described by the method in~\cite{Horn87closed-formsolution} and consider $K_{loop}$ as the loop keyframe of $K_t$ if we find $T_{sim}$ with enough inliers (more than 100).

\subsubsection{Loop Merging}
Loop merging corrects any existing drift in the estimations of $K_t$ and its neighbouring keyframes. We update the pose for $K_t$ and its neighbouring keyframes using $T_{sim}$. There is a set of 3D points $M_{loop}$ that are visible to $K_{loop}$ and its neighbours whereas the set of 3D points $M_t$ that are visible to $K_t$ and its neighbours. We project the 3D points belong to $M_{loop}$ to the current keyframe $K_t$ and its neighbours and check for the 3D-2D correspondences between these projected points and the 3D points belong to $M_t$. We merge all those map points where the 3D-2D correspondences are found and those that were inliers in $T_{sim}$. Global bundle adjustment further refines the poses of all keyframes in the entire loop and corresponding map points.

\begin{figure}
\begin{center}
\includegraphics[width=1\linewidth]{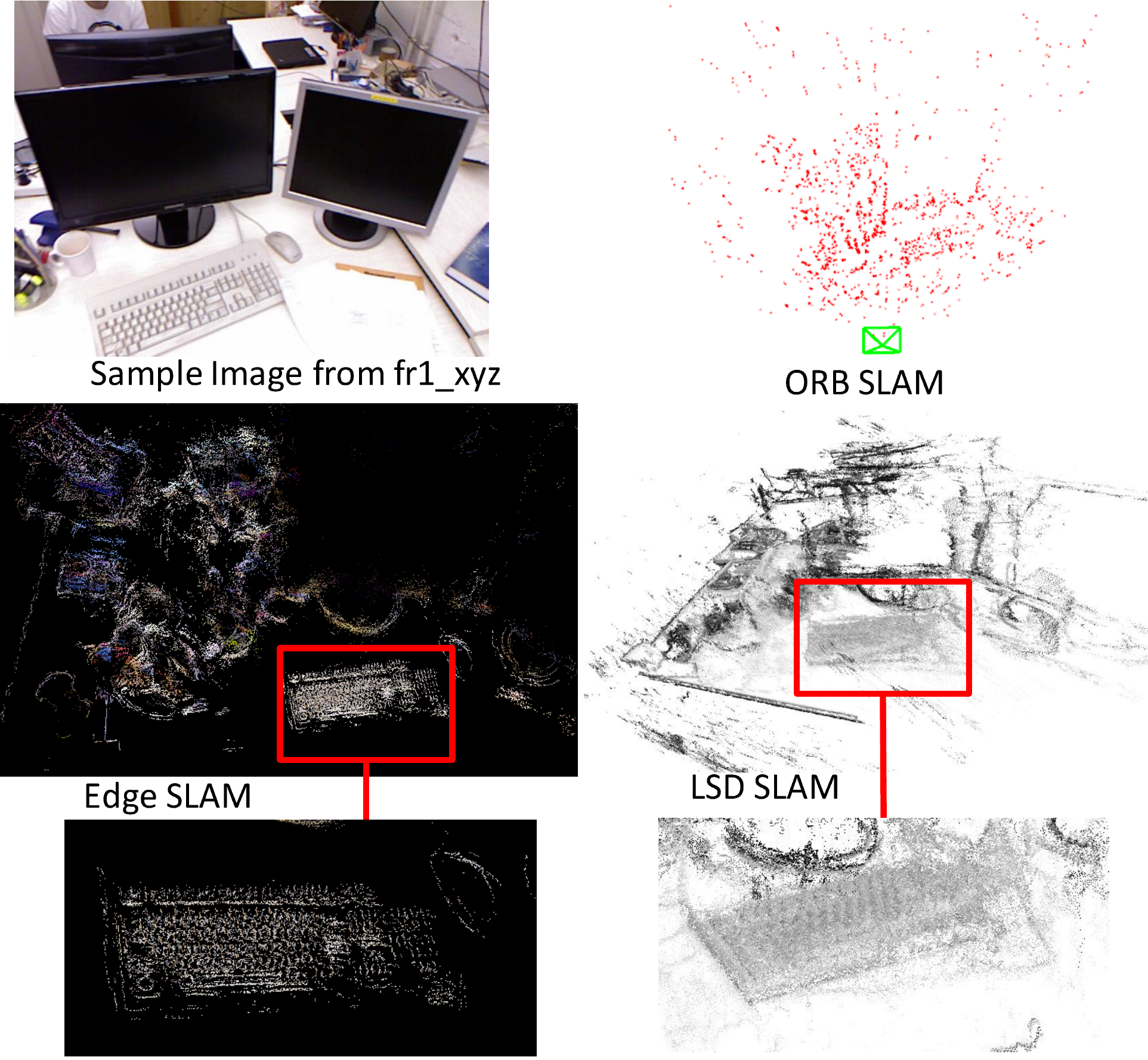}
\end{center}
\caption{The figure shows comparisons among reconstructed structures by ORB SLAM, LSD SLAM and our Edge SLAM on $fr1\_xyz$ \cite{sturm12iros} sequence. ORB SLAM structure is too sparse to understand the environment. LSD SLAM generates a semi-dense structure, where multiple overlapped structures are visible. Our Edge SLAM generates a 3D structure which yield semantic understanding of the environment.}
\label{fig:strucCompare}
\end{figure}

\section{Experimental Results}
\label{sec:Result}
We have used an Intel Core i7-7700 (4 cores @ 3.60GHz) with 8Gb RAM, for implementation of our SLAM pipeline. We extensively experiment with TUM RGB-D benchmark datasets \cite{sturm12iros} and ICL-NUIM \cite{handa:etal:ICRA2014} dataset. We have used TUM RGBD Benchmarking tool \cite{sturm12iros} to compare the camera estimations by our Edge SLAM against ground-truth. Table~\ref{tab:compare} also shows the comparison of Absolute Trajectory Errors by our Edge SLAM, LSD-SLAM, ORB-SLAM, Edge VO and PL SLAM \cite{pumarola2017plslam} where it is evident that unlike the existing pipelines, our Edge SLAM works on all kind of datasets reliably. Moreover, we like to emphasize that our pipeline also produces accurate structure and camera estimations even for a low-textured environment where most of the existing SLAMs fail in tracking (six rows from bottom of Table~\ref{tab:compare}). Most recent visual odometry pipeline Edge VO is limited to produce only camera localization without any map and produces the most erroneous result as shown on Table~\ref{tab:compare}. An example on loop closing is presented on $fr2\_desk$ \cite{sturm12iros} sequence in figure~\ref{fig:loop} where the RMS of ATE is reduced from 9.5 cm to 1.75 cm after loop closing.

\begin{table}[t]
\begin{center}
\resizebox{\columnwidth}{!}{%
\begin{tabular}{|l|c|c|c|c|c|}
\hline
\multicolumn{6}{|c|}{Absolute keyframe Trajectory RMS Error (cm)} \\
 \hline
 Sequences & Edge SLAM & ORB-SLAM & LSD-SLAM & Edge VO & PL SLAM\\
 \hline\hline
 $fr1\_xyz$ & 1.31 & {\bf 0.90} & 9.00 & 16.51 & 1.21\\
 $fr2\_desk$ & 1.75 & {\bf 0.88} & 4.57 & 33.67 & -\\
 $fr2\_xyz$ & 0.49 & {\bf 0.30} & 2.15 & 21.41 & 0.43\\
 $fr3\_str\_tex\_near$ & {\bf 1.12} & 1.58 & \textcolor{red}{X} & 47.63 & 1.25\\
 $fr3\_str\_tex\_far$ & {\bf 0.65} & 0.77 & 7.95 & 121.00 & 0.89\\
 $fr3\_str\_notex\_near$ & {\bf 8.29} & \textcolor{red}{X} & \textcolor{red}{X} & 101.03 & -\\
 $fr3\_str\_notex\_far$ & {\bf 6.71} & \textcolor{red}{X} & \textcolor{red}{X} & 41.76 & -\\
 $ICL/office0$ & {\bf 3.21} & 5.67 & \textcolor{red}{X} & \textcolor{red}{X} & -\\
 $ICL/office1$ & {\bf 19.5} & \textcolor{red}{X} & \textcolor{red}{X} & \textcolor{red}{X} & -\\
 $ICL/office2$ & {\bf 2.97} & 3.75 & \textcolor{red}{X} & \textcolor{red}{X} & -\\
 $ICL/office3$ & {\bf 4.58} & 16.18 & \textcolor{red}{X} & \textcolor{red}{X} & -\\
 \hline
\end{tabular}}
\end{center}
\caption{keyframe localization error comparison. \textcolor{red}{X} denote unsuccessful cases.}
\label{tab:compare}
\end{table}

\begin{table}[t]
\begin{center}
\resizebox{0.6\columnwidth}{!}{%
\begin{tabular}{|c|c|}
 \hline
 \bf{Method} & \bf{Mean (ms)}\\
 \hline
 Edge extraction & 13\\
 \hline
 Correspondence generation & 7\\
 \hline
 Keyframe selection & 14\\
 \hline
 Pose estimation & 11\\
 \hline
 Map generation & 5 \\
 \hline
 Local bundle adjustment (on 5 keyframes) & 175 \\
 \hline 
\end{tabular}}
\end{center}
\caption{Module wise mean execution time on $fr2\_desk$ \cite{sturm12iros}.}
\label{tab:timeanalysis}
\end{table}

In addition to better camera localization, our pipeline produces significantly improved structure compared with existing SLAMs. Figure~\ref{fig:strucCompare} shows an example of structure comparison of our Edge SLAM with LSD \& ORB SLAM where, it is evident that our pipeline produce superior quality structure against existing SLAM pipelines. We left out the structure comparison with Edge VO as the pipeline intended to produce only camera motion. We also present the module wise time analysis in Table~\ref{tab:timeanalysis} calculated on $fr2\_desk$ \cite{sturm12iros} sequence. Our camera tracking method runs on 17 fps, where as the mapping method runs on 4 fps using an unoptimized implementation.

\section{Conclusion}
We present a visual SLAM pipeline with the focus on track in both textured as well as very low-textured environments and building recognizable maps. We start with initializing SLAM through a validation process that produces better initialization compared with state-of-the-art visual SLAMs. We present a novel local optimization method for stable camera estimation in the situations where camera tracking becomes unreliable in a very low-textured challenging environment. Our pipeline is capable of an efficient and reliable loop closing using structural properties of edges in images. The pipeline shows a significant improvement in map generation in terms of semantic understanding. 

{\small
\bibliographystyle{ieee}
\bibliography{egbib}
}

\end{document}